\documentclass{tlp}

\usepackage{amsmath}
\usepackage{graphicx}
\usepackage[english]{babel}

\usepackage{multirow}
\usepackage{times}
\usepackage{helvet,courier}
\usepackage{courier,txfonts}
\DeclareMathAlphabet{\mathitbf}{OML}{cmm}{b}{it}   

\begin{document}

\lefttitle{Vaishak Belle}

\jnlPage{1}{8}
\jnlDoiYr{2021}
\doival{10.1017/xxxxx}

\title[Theory and Practice of Logic Programming]{Toward A Logical Theory Of Fairness and Bias}

\begin{authgrp}
\author{\sn{Vaishak} \gn{Belle}}
\affiliation{University of Edinburgh \& Alan Turing Institute, UK}
\end{authgrp}
\let\cite\citep 
\history{\sub{xx xx xxxx;} \rev{xx xx xxxx;} \acc{xx xx xxxx}}

\maketitle

%
%
%
\newcommand{\bl}{\ti{Blocked}}
\newcommand{\walle}{\textsc{WallFar}}
\newcommand{\walla}{\textsc{WallClose}}
\newcommand{\golog}{\textsc{Golog}{}}
\newcommand{\alt}{\ti{Alt}}
\newcommand{\gap}{\smallskip}
\newcommand{\xmove}{\textit{move}}
\newcommand{\chop}{\emph{chop}}

\renewcommand{\sub}{_}
\newcommand{\xs}{x^*}
\newcommand{\ys}{y^*}
\newcommand{\thereis}{\exists}
\newcommand{\rint}{\int_{x \in \real}}
\newcommand{\is}{\thereis}
\def\su{^}           
\newcommand{\rational}{\rat}
\newcommand{\vecx}{{\vec x}}

\newcommand{\xO}{{x_1}}
\newcommand{\xT}{{x_2}}
\newcommand{ \xo}{\xO}
\newcommand{ \xt}{\xT}

\newcommand{\rationals}{\rats}
\newcommand{\no}{\noindent}
\newcommand{\rarrow}{\rightarrow}
\newcommand{\oisame}{\approx\sub{alt}}
\newcommand{\rhs}{\textsc{RHS}}
\newcommand{\larrow}{\leftarrow}

\newcommand{\real}{{\mathbb{R}}}
\newcommand{\rat}{{\mathbb{Q}}}
\newcommand{\rats}{{\rat\sub{[0,1]}}}
\newcommand{\nat}{{\mathbb{N}}}
\newcommand{\dom}{\textit{dom}}
\newcommand{\sig}{\textit{sig}}
\newcommand{\sat}{\models}
\newcommand{\dtur}{\models}
\newcommand{\infers}{\vdash}
\newcommand{\stur}{\vdash}
\newcommand{\rimp}{\Rightarrow}
\newcommand{\limp}{\Leftarrow}
\newcommand{\dimp}{\Leftrightarrow}
\newcommand{\bor}{\bigvee}
\newcommand{\dummy}[1]{\textrm{d}{#1}}

\newcommand{\band}{\bigwedge}
\newcommand{\union}{\cup}
\newcommand{\inter}{\cap}
\newcommand{\xx}{{\bf x}}
\newcommand{\yy}{{\bf y}}
\newcommand{\uu}{{\bf u}}
\newcommand{\FF}{{\bf F}}
\newcommand{\nn}{{n_1,\ldots,n_k}}
\newcommand{\disp}{\displaystyle}
\newcommand{\dis}{\disp}
\newcommand{\natnum}{{\sl N}}

\newcommand{\lt}{<}
\newcommand{\gt}{>}
\newcommand{\all}{\forall}
\newcommand{\infinity}{\infty}
\newcommand{\kb}{\D\sub 0}
\newcommand{\bigfootnote}[1]{{\footnote{\normalsize #1}}}
\newcommand{\medfootnote}[1]{{\footnote{\small #1}}}
\newcommand{\theory}{\Sigma}
\newcommand{\theorydyn}{\theory\sub {\it dyn}}
\newcommand{\theoryinit}{\theory\sub 0}
\newcommand{\theoryinits}{\theory\sub 0 '}
\newcommand{\bd}{\bf}
\newcommand{\ori}{\ti{rotate}}
\newcommand{\laser}{\ti{laser}}
\newcommand{\ymove}{\textit{up}}
\newcommand{\near}{\ti{NearestLeft}}
\newcommand{\tow}{\emph{towards}}
\newcommand{\away}{\emph{away}}
\newcommand{\up}{\textit{up}}
\newcommand{\nup}{\textit{up}}
\newcommand{\nfwd}{\nup}
\newcommand{\nmove}{\nup}
\newcommand{\ypos}{\textit{v}}
\newcommand{\ask}{\textit{ask}}
\newcommand{\broad}{\textit{inform}}
\newcommand{\ann}{\textit{announce}}
\newcommand{\noop}{\emph{null}}
\newcommand{ \move}{\xmove}
\newcommand{\sol}{\ti{Solid}}

\newcommand{\bbox}{\vrule height7pt width4pt depth1pt}
\newcommand{\eprf}{\bbox\vspace{0.1in}}

\newcommand{\imove}{\ti{imove}}
\newcommand{\sumF}{\textit{sum}}
\newcommand{\fullintegral}{{\int_{-\infty}^{\infty}}}
\newcommand{\fullintegrala}{{\int_{\real^m}}}
\newcommand{\fullintegralx}{\int_{\vec x}}
\newcommand{ \fullx}{\fullintegralx}
\newcommand{\fullsumy}{\sum_{\vec y}}
\newcommand{\stara}{\textrm{\upshape{P3}}}
\newcommand{\imp}{\Rightarrow}

\newcommand{\A}{{\cal A}}
\newcommand{\B}{{\cal B}}
\newcommand{\D}{\theory}
\newcommand{\E}{\textit{E}}
\newcommand{\F}{{\cal F}}
\newcommand{\I}{{\cal I}}
\newcommand{\J}{{\cal J}}
\newcommand{\K}{{\cal K}}
\renewcommand{\L}{{\cal L}}
\newcommand{\M}{{\cal M}}
\newcommand{\N}{{\cal N}}
\newcommand{\prims}{\textit{Prims}}
\newcommand{\s}{\sigma}
\newcommand{\so}{\s_0}
\newcommand{\obs}{\textit{SF}}
\newcommand{\la}{\langle}
\newcommand{\ra}{\rangle}
\newcommand{\fsa}{{\Sigma}}
\newcommand{\raw}{\rarrow}
\newcommand{\goto}{\raw}
\newcommand{\done}{\textit{stop}}
\newcommand{\eps}{\epsilon}
\newcommand{\qinit}{{Q_0}}
\newcommand{\qi}{{q\sub 0}}
\newcommand{\qf}{{q\sub F}}
\newcommand{\wi}{{s\sub 0 }}
\newcommand{\prob}{l}

\newcommand{\apply}{\emph{end}}
\newcommand{\last}{\emph{last}}

\newcommand{\en}{\apply}
\newcommand{\bibtext}{Notes}
\newcommand{\weak}{{\textsc{\textbf{ONE}}}}
\newcommand{\pc}{\textsc{\textbf{PC}}}
\newcommand{\ter}{\textsc{\textbf{TER}}}
\newcommand{\lter}{\textsc{\textbf{LTER}}}
\newcommand{\BEL}{{\textsc{\textbf{BEL}}_\kappa}}
\newcommand{\MOST}{{\textsc{\textbf{GEQ}}_\kappa}}
\newcommand{\eqs}{\textsc{=1}}
\newcommand{\lpc}{\textsc{\textbf{LPC}}}
\newcommand{\lexpt}{\textsc{\textbf{LCOST}}}
\newcommand{\lexptt}{\textsc{\textbf{LPCOST}}}
\newcommand{\exptt}{\lexptt}

\newcommand{\boundcorr}{\textsc{\textbf{BND}}}

\newcommand{\cprt}{\textsc{CPRT}}
\newcommand{\lin}{\textsc{LL}}
\newcommand{\mdp}{\textsc{MDP}}
\newcommand{\hg}{\textsc{HD}}
\newcommand{\acyc}{\textsc{\textbf{ACYC}}}
\renewcommand{\ss}{{s^*}}

\newcommand{\calO}{{\cal O}}
\newcommand{\Ocal}{{\cal O}}
\newcommand{\Hcal}{{\cal H}}
\renewcommand{\P}{{\cal P}}
\newcommand{\Q}{{\cal Q}}
\newcommand{\R}{{\cal R}}
\renewcommand{\S}{{\cal S}}
\newcommand{\T}{{\cal T}}
\newcommand{\V}{{\cal V}}
\newcommand{\W}{{\cal W}}
\newcommand{\X}{{\cal X}}
\newcommand{\Y}{{\cal Y}}
\newcommand{\Z}{{\cal Z}}

\newcommand{\Kone}{{\cal K}_1}
\newcommand{\abs}[1]{\left| #1\right|}
\newcommand{\set}[1]{\left\{ #1 \right\}}
\newcommand{\Ki}{{\cal K}_i}
\newcommand{\Kn}{{\cal K}_n}
\newcommand{\st}{\, \vert \,} 
\newcommand{\stc}{\, : \,} 
\newcommand{\lan}{\langle}
\newcommand{\like}{\ti{Err}}
\newcommand{\dd}{\displaystyle}
\newcommand{\calL}{{\cal L}}
\newcommand{\natt}{\textit{Natural}}
\newcommand{\summ}{\textsc{SUM}}

\newcommand{\ran}{\rangle}
\newcommand{\<}{\langle}
\renewcommand{\>}{\rangle}
\newcommand{\pro}{\textit{Pro}}
\newcommand{\lang}{\textsc{Allegro}}
\newcommand{\eval}{\textsc{eval}}
\newcommand{\interp}{\textsc{int}}
\newcommand{\prego}{\textsc{prego}}
\newcommand{\trans}{\emph{Trans}}
\newcommand{\final}{\emph{Final}}
\newcommand{\inip}{\Theta}
\newcommand{\prog}{\textsc{prog}}
\newcommand{\sits}{\textsc{sits}}
\newcommand{\scheme}{\textsc{scheme}}
\newcommand{\racket}{\textsc{racket}}
\newcommand{\conf}{\textit{Conf}}
\newcommand{\expt}{\textit{Exp}}
\newcommand{\di}{\displaystyle}
\newcommand{\tool}{\textsc{I}}
\newcommand{\xpostt}{\texttt{h}}
\newcommand{\hh}{\xpostt}

\newcommand{\expval}{\textsc{expvalue}}

\newcommand{\dtg}{\textsc{dtgolog}}

\newcommand{\dn}{\G}

\newtheorem{nlem}{Lemma}   
\newtheorem{Ob}{Observation}
\newtheorem{pps}{Proposition}       
\newcommand{\poss}{{\it Poss}}
\renewcommand{\sf}{{\it SF}}
\newcommand{\exec}{{\it Exec}}
\newtheorem{defn}{Definition}
\newtheorem{crl}{Corollary}
\newtheorem{cl}{Claim}
\newcommand{\pf}{\par\noindent{\bf Proof}~~}
\newcommand{\eg}{\emph{e.g.,}~}
\newcommand{\ie}{\emph{i.e.,}~}
\newcommand{\cf}{cf.~}
\newcommand{\etal}{et al.\ }
\newcommand{\resp}{resp.\ }
\newcommand{\respc}{resp.,\ }
\newcommand{\wrt}{with respect to~}
\newcommand{\re}{r.e.}
\newcommand{\If}{{\bf if}}

\newcommand{\Then}{{\bf then}}
\newcommand{\Until}{{\bf until}}
\newcommand{\Else}{{\bf else}}
\newcommand{\Repeat}{{\bf repeat}}
\newcommand{\cA}{{\cal A}}
\newcommand{\cE}{{\cal E}}
\newcommand{\cF}{{\cal F}}
\newcommand{\cI}{{\cal I}}
\newcommand{\ti}{}

\newcommand{\now}{now}
\newcommand{\ifc}{\textsc{{If}}}  
\newcommand{\error}{\ti{Err}}
\newcommand{\thenc}{\textsc{{Then}}}
\newcommand{\elsec}{\textsc{{Else}}}  
\newcommand{\probabbr}[3]{\lan #1.#2\to #3 \ran}
\newcommand{\dens}{Density}
\newcommand{\func}{P}
\newcommand{\ivar}{\iota}
\newcommand{\sone}{S\sub 1}
\newcommand{\sstar}{{s^*}}
\newcommand{\splus}{{s^+}}
\newcommand{\spplus}{{s^{++}}}
\newcommand{\sbullet}{{s^\bullet}}
\newcommand{\init}{\textit{Init}}
\newcommand{\initd}{{\D\sub 0}}

\newcommand{\xsensor}{\textit{xsonar}}
\newcommand{\ysensor}{\textit{ysonar}}
\newcommand{\xsense}{\textit{sonar}}
\newcommand{\sonar}{\xsense}
\newcommand{\defeq}{\doteq}

\newcommand{\cN}{{\cal N}}
\newcommand{\cR}{{\cal R}}
\newcommand{\cS}{{\cal S}}
\newcommand{\xpos}{h}
\newcommand{\forget}{{\it Forget}}
\newcommand{\true}{\mbox{\it true}}
\newcommand{\false}{\mbox{\it false}}

\newcommand{\oknow}{\mathitbf{O}}
\newcommand{\Bel}{\mathitbf{B}}
\newcommand{\know}{\mathitbf{K}}
\newcommand{\wh}{\mathitbf{W}}
\newcommand{\knows}{Knows}
\newcommand{\kwhether}{KWhether}
\newcommand{\kref}{KRef}
\newcommand{\modalK}{\know}
\newcommand{\bhlbel}{Bel}
\newcommand{\bel}{\bhlbel}

\newcommand{\ins}{{S}\sub  {0}}

\newcommand{\es}{{\cal E}{\cal S}}
\newcommand{\ol}{{\cal O}{\cal L}}

\newcommand{\normal}{\textsc{Norm}}
\newcommand{\norm}{\normal}
\newcommand{\normdenom}{\frac{1}{\gamma}}
\newcommand{\equal}{\textsc{Equal}}
\newcommand{\bound}{\textsc{Bound}}

\renewcommand{\emptyset}{\set{}}

\begin{abstract}
Fairness in machine learning is of considerable interest in  recent years owing to the propensity of algorithms trained on historical data to amplify and perpetuate historical biases. In this paper, we argue for a formal reconstruction of fairness definitions, not so much to replace existing definitions but to ground their application in an epistemic setting and allow for rich environmental modelling. Consequently we look into three notions: fairness through unawareness, demographic parity and counterfactual fairness, and formalise these in the epistemic situation calculus.
\end{abstract}

\begin{keywords}
Logic, Fairness, Bias, Situation Calculus, Knowledge, Action
\end{keywords}

\hypertarget{introduction}{%
\section{Introduction}\label{introduction}}

Machine Learning techniques have become pervasive across a range of
different applications, and are the source of considerable excitement
but also debate. For example, they are now widely used in areas as
disparate as recidivism prediction, consumer credit-risk analysis and
insurance pricing \citep{Recidivism_study, ML_for_credit}. In some of
these applications, the prevalence of machine learning techniques has
raised concerns about the potential for learned algorithms to become
biased against certain groups. This issue is of particular concern in
cases when algorithms are used to make decisions that could have
far-reaching consequences for individuals (for example in recidivism
prediction) \citep{Recidivism_study, ProPublica}. Attributes which the
algorithm should be ``fair'' with respect to are typically referred to
as \emph{protected} attributes. The values to these are often hidden
from the view of the decision maker (whether automated or human). There
are multiple different potential fields that might qualify as protected
attributes in a given situation, including ethnicity, sex, age,
nationality and marital status \citep{Learning_Fair_Reps}. Ideally, such
attributes should not affect any prediction made by ``fair'' algorithms.
However, even in cases where it is clear which attributes should be
protected, there are multiple (and often mutually exclusive) definitions
of what it means for an algorithm to be unbiased with respect to these
attributes, and there is disagreement within the academic community on
what is most appropriate
\citep{Fairness_through_awareness, Counterfactual_Fairness, Disparate_Impact_&_Historical}.

However, even amid pressing concerns that algorithms currently in use
may exhibit racial biases, there remains a lack of agreement about how
to effectively implement fairness, given the complex socio-technical
situations that such applications are deployed in and the background
knowledge and context needed to assess the impact of outcomes (e.g.,
denying a loan to someone in need).

To address such issues broadly, an interesting argument has been
championed by the symbolic community: by assuming a rich enough
understanding of the application domain, we can encode machine ethics in
a formal language. Of course, with recent advances in statistical
relational learning, neuro-symbolic AI and inductive logic programming
\citep{raedt2016statistical,muggleton2012ilp}, it is possible to
integrate low-level pattern recognition based on sensory data with
high-level formal specifications. For example, the \emph{Hera} project
\citep{lindner2017hera} allows for the implementation of several kinds
of (rule-based) moral theory to be captured. \emph{Geneth}
\citep{anderson2014geneth} uses inductive logic programming to create
generalised moral principles from the judgements of ethicists about
particular ethical dilemmas, with the system's performance being
evaluated using an \emph{ethical Turing test}. On the formalisation
side, study of moral concepts has long been a favored topic in the
knowledge representation community
\citep{conway2013deontological,sep-ethics-deontological,czelakowski1997action,hooker2018toward},
that can be further coupled against notions of beliefs, desires and
intentions \citep{broersen2001boid,georgeff1998belief}. Finally, closer
to the thrust of this paper, \citep{pagnucco2021epistemic} formalize
consequentialist and deontological ethical principles in terms of
``desirable'' states in the epistemic situation calculus, and
\citep{classen2020dyadic} formalize obligations using situation calculus
programs.

\hypertarget{contributions}{%
\section{Contributions}\label{contributions}}

Our thesis, in essence, is this: complementing the vibrant work in the
ML community, it is worthwhile to study ethical notions in formal
languages. This serves three broad objectives:

\begin{enumerate}
\item
  We can identify what the system needs to know versus what is simply
  true \citep{DBLP:journals/tocl/Reiter01,DBLP:journals/jair/HalpernM14}
  and better articulate how this knowledge should impact the agent's
  choices. It is worth remarking that epistemic logic has served as the
  foundation for investigating the impact of knowledge on plans and
  protocols
  \citep{DBLP:conf/aaai/Levesque96,DBLP:journals/sLogica/LesperanceLLS00,DBLP:conf/tark/HalpernPR09}.
\item
  We implicitly understand that we can further condition actions against
  background knowledge (such as ontologies and databases), as well as
  notions such as intentions and obligations \citep{Sardina:2010aa}.
\item
  We can position the system's actions not simply as a single-shot
  decision or prediction, as is usual in the ML literature, but as a
  sequence of complex events that depend on observations and can involve
  loops and recursion: that is, in the form of programs
  \citep{Levesque97-Golog}.
\end{enumerate}

It would beyond the scope of a single paper to illustrate the interplay
between the three objectives except in some particular application
scenario. Thus, we focus on the interplay between A and C in the sense
of advocating a ``research agenda,'' rather than a single technical
result, or a demonstration of a single application. In particular, what
we seek to do is a formal reconstruction of some fairness definitions,
not so much to replace existing definitions but to ground their
application in an epistemic, dynamic setting. Consequently we look into
three notions: fairness through unawareness, demographic parity and
counterfactual fairness, and formalise these in the epistemic situation
calculus \citep{citeulike:528170,Lakemeyer:2011:SCU:1897346.1897552}. In
particular, our contributions are as follows:

\begin{itemize}
\item
  Consider the notion of fairness through unawareness (FTU) in machine
  learning. Here, a ``fair" classifier is one that predicts outputs by
  not using any information about protected attributes. In a dynamic
  setting, imagine a (virtual or physical) robot that is acting in
  service of some objective \(\phi.\) For example, in a loan setting,
  which is classically treated as a static model in machine learning, we
  can expect intelligent automated agents to carry out many operations:
  check the yearly budget of the bank to determine the total amount to
  be loaned, rank applicants based on risk, determine the impact of
  recession, and ultimately synthesize a plan to achieve \(\phi\) (loan
  approval), but by virtue of FTU, it should never be the case that the
  agent has had access to protected information. In this paper, we
  provide a simple but general definition to capture that idea, in a
  manner that distinguishes what is true from what is known by the
  agent.
\item
  Analogously, consider the notion of demographic parity (DP). It is
  understood as a classifier that is equally likely to make a positive
  prediction regardless of the value of the protected attribute. For
  example, the proportion of men who are granted loans equals the
  proportion of women granted loans. So, if \(\phi(x)\) is the granting
  of a loan to individual \(x,\) how do we capture the notion that the
  agent has synthesized a plan that achieves \(\phi(x)\) for both males
  as well as females? What would it look like for planning agents that
  want to conform to both FTU and DP? What if, instead of DP, we wished
  to only look at those granted loans, and among this group, we did not
  want the classifier to discriminate based on the individual's gender?
  For all these cases, we provide definitions in terms of the agent's
  mental state and action sequences that the agent knows will achieve
  \(\phi(x)\) \cite{DBLP:conf/aaai/Levesque96}.
\item
  Finally, counterfactual fairness insists that the prediction should
  not differ if the individual's protected attributes take on a
  different value. For a planning agent to ensure this, we would need to
  make sure that \emph{deleting} facts about the current value for an
  individual \(x\)'s protected attribute and \emph{adding} a different
  value still achieves \(\phi(x)\) after the sequence. We characterize
  this using the notion of \emph{forgetting} because we permit, in
  general, any arbitrary first-order theory for the initial knowledge
  base, and not just a database interpreted under the closed-world
  assumption.
\end{itemize}

These definitions can be seen to realize a specification for ``fair"
cognitive robots: that is, reasoning and planning agents
\cite{lakemeyer2007chapter} that ensure through the course of their
acting that, say, they never gain knowledge about the protected
attributes of individuals, and guarantee that individuals are not
discriminated based on values to these attributes.

It should be clear that our definitions are loosely inspired by the ML
notions. And so our formalisation do not argue for one definition over
another, nor challenge any existing definition. We do, however, believe
that studying the effects of these definitions in a dynamic setting
provides a richer context to evaluate their appropriateness. Moreover, a
formalisation such as ours lends itself to various types of
implementations. For example, the synthesis of (epistemic) programs and
plans
\cite{wang2005nested,baral_et_al:DR:2017:8285,muise-aaai-15,Clasen:2008ly,DBLP:conf/kr/McIlraithS02}
that achieve goals in socio-technical applications in a fair manner is
an worthwhile research agenda. Likewise, enforcing fairness constraints
while factoring for the relationships between individuals in social
networks \cite{farnadi2018fairness}, or otherwise contextualising
attributes against other concepts in a relational knowledge base
\cite{aziz2018knowledge,fu2020fairness} are also worthwhile. By
stipulating an account in quantified logic, it becomes possible to
further unify such proposals in a dynamic setting.

\textbf{Logic and fairness.} Let us briefly remark on closely related
efforts. At the outset, note that although there has been considerable
work on formalizing moral rules, there is no work (as far as we are
aware) on the formalization of fairness and bias in a \emph{dynamic
epistemic} setting, where we need to explicate the interaction between
actions, plans and meta-beliefs. However, there is some work that
tackles epistemic and logical aspects.

For example, the work of \cite{yam} considers a statistical epistemic
logic and its use for the formalisation of statistical accuracy as well
as fairness, including the criterion of equality of opportunity. There
are a few key differences to our work: that work is motivated by a
probabilistic reconstruction of prediction systems by appealing to
distance measures, and so knowledge is defined in terms of accessibility
between worlds that are close enough. The language, moreover, allows for
``measurement" variables that are interpreted statistically. In
contrast, our account is not (yet) probabilistic, and if our account
were to be extended in that fashion, the most obvious version would
reason about degrees of belief \cite{Bacchus1999171,Belle2017ac}; see
\cite{bacchus1996statistical} for discussions on the differences between
statistical belief and degrees of belief. Moreover, our account is
dynamic, allowing for explicit modalities operators for actions and
programs. Consequently, our definitions are about studying how, say, the
agent remains ignorant about protected attributes when executing a plan.

Be that as it may, the work of \cite{yam} leads to an account where
fairness can be expressed as a logical property using predicates for
protected attributes, remarkably similar in spirit to our approach if
one were to ignore actions. This should, in the very least, suggest that
such attempts are very promising, and for the future, it would be
worthwhile to conduct a deeper investigation on how these formalisation
attempts can be synthesized to obtain a general probabilistic logical
account that combines the strength of dynamic epistemic languages and
statistical measures. (In a related vein to \cite{yam},
\citep{liu2022logic} seek to axiomatize ML sytems for the purpose of
explanations in a modal logic.) An entirely complementary effort is the
use of logic for verifying fair models \citep{ignatiev2020towards},
where existing definitions and classifiers are encoded using logical
functions and satisfiability modulo theories.

To summarize, all these differ from our work in that we are attempting
to understand the interplay between bias, action and knowledge, and not
really interested in capturing classifiers as objects in our language.
Thus, our work, as discussed above, can be seen as setting the stage for
\emph{``fair'' cognitive robots}. There is benefit to unifying these
streams, which we leave to the future.

\hypertarget{sec:reconstructing_the_epistemic_situation_calculus}{%
\section{A logic for knowledge and
action}\label{sec:reconstructing_the_epistemic_situation_calculus}}

We now introduce the logic \(\es\)
\citep{LakemeyerLevesque2004}.\footnote{Our choice of language may seem
  unusual, but it is worth noting that this language is a modal
  syntactic variant of the classical epistemic situation that is better
  geared for reasoning about knowledge
  \citep{Lakemeyer:2011:SCU:1897346.1897552}. But more importantly, it
  can be shown that reasoning about actions and knowledge reduces to
  first-order reasoning via the so-called regression and representation
  theorems \citep{LakemeyerLevesque2004}. (For space reasons, we do not
  discuss such matters further here.) There are, of course, many works
  explicating the links between the situation calculus and logic
  programming; see, for example, \citep{lee2012reformulating}. See also
  works that link the situation calculus to planning, such as
  \cite{Clasen:2008ly,belle2022analyzing,sardina2004semantics,DBLP:conf/aips/BaierFM07}.}
The non-modal fragment of \(\es\) consists of standard first-order logic
with \(=\). That is, connectives \(\set{\land, \forall, \neg}\),
syntactic abbreviations \(\set{\exists,\equiv, \supset}\) defined from
those connectives, and a supply of variables variables
\(\set{x,y,\ldots, u, v, \ldots}\). Different to the standard syntax,
however, is the inclusion of (countably many) \emph{standard names} (or
simply, names) for both objects and actions \(\R,\) which will allow a
simple, substitutional interpretation for \(\forall\) and \(\exists.\)
These can be thought of as special extra constants that satisfy the
unique name assumption and an infinitary version of domain closure.

Like in the situation calculus, to model immutable properties, we assume
rigid predicates and functions, such as {\emph{IsPlant(x)}} and
{\emph{father(x)}} respectively. To model changing properties, \(\es\)
includes fluent predicates and functions of every arity, such as
{\emph{Broken(x)}} and {\emph{height(x)}}. Note that there is no longer
a situation term as an argument in these symbols to distinguish the
fluents from the rigids. For example, \(\es\) also includes
distinguished fluent predicates \(\poss\) and \(\sf{}\) to model the
executability of actions and capture sensing outcomes respectively, but
they are unary predicates (that is, in contrast to the classical
situation calculus \cite{reiter2001knowledge} because they no longer
include situation terms.) Terms and formulas are constructed as usual.
The set of ground atoms \(\P\) are obtained, as usual, from names and
predicates.

There are four modal operators in \(\es\): \([a], \Box, \know\) and
\(\oknow.\) For any formula \(\alpha,\) we read
\([a]\alpha, \Box\alpha\) and \(\know\alpha\) as ``\(\alpha\) holds
after \(a\)", ``\(\alpha\) holds after any sequence of actions" and
``\(\alpha\) is known,'' respectively. Moreover, \(\oknow\alpha\) is to
be read as ``\(\alpha\) is only-known.'' Given a sequence
\(\delta = a\sub 1 \cdots a\sub k,\) we write \([\delta]\alpha\) to mean
\([a\sub 1] \cdots [a\sub k]\alpha.\)

In classical situation calculus parlance, we would use \([a]\alpha\) to
capture successor situations as properties that are true after an action
in terms of the current state of affairs. Together with the \(\Box\)
modality, which allows to capture quantification over situations and
histories, basic action theories can be defined. Like in the classical
approach, one is interested in the entailments of the basic action
theory.

\textbf{Semantics.} Recall that in the simplest setup of the
possible-worlds semantics, worlds mapped propositions to \(\set{0,1}\),
capturing the (current) state of affairs. \(\es\) is based on the very
same idea, but extended to dynamical systems. So, suppose a world maps
\(\P\) and \(\Z\) to \(\set{0,1}\).\footnote{We need to extend the
  mapping to additionally interpret fluent functions and rigid symbols,
  omitted here for simplicity.} Here, \(\Z\) is the set of all finite
sequences of action names, including the empty sequence \(\lan\ran.\)
Let \(\W\) be the set of all worlds, and \(e\subseteq \W\) be the
\emph{epistemic state}. By a \emph{model}, we mean a triple \((e,w,z)\)
where \(z\in \Z.\) Intuitively, each world can be thought of as a
situation calculus tree, denoting the properties true initially but also
after every sequence of actions. \(\W\) is then the set of all such
trees. Given a triple \((e,w,z)\), \(w\) denotes the real world, and
\(z\) the actions executed so far.

To account for how knowledge changes after (noise-free) sensing, one
defines \(w' \sim\sub z w\), which is to be read as saying ``\(w'\) and
\(w\) agree on the sensing for \(z\)'', as follows:

\begin{itemize}
\item
  if \(z=\lan\ran,\) \(w' \sim\sub {z} w\) for every \(w'\); and
\item
  \(w'\sim \sub {z\cdot a} w\) iff \(w'\sim\sub {z} w,\)
  \(w'[\poss(a),z]=1\) and \(w'[\sf(a),z] = w[\sf(a),z].\)
\end{itemize}

This is saying that initially, we would consider all worlds compatible,
but after actions, we would need the world \(w'\) to agree on the
executability of actions performed so far as well as agree on sensing
outcomes. The reader might notice that this is clearly a reworking of
the successor state axiom for the knowledge fluent in
\citep{citeulike:528170}.

With this, we get a simply account for truth. We define the satisfaction
of formulas wrt (with respect to) the triple \((e,w,z)\), and the
semantics is defined inductively:

\begin{itemize}
\item
  \(e,w,z\models p\) iff \(p\) is an atom and \(w[p,z] =1\);
\item
  \(e,w,z \models \alpha\land\beta\) iff \(e,w,z\models \alpha\) and
  \(e,w,z\models \beta;\)
\item
  \(e,w,z\models \neg\alpha\) iff \(e,w,z\not\models \alpha;\)
\item
  \(e,w,z\models \forall x\alpha\) iff \(e,w,z\models \alpha^x_n\) for
  all \(n \in \R;\)
\item
  \(e,w,z\models [a]\alpha\) iff \(e,w,z\cdot a\models \alpha;\)
\item
  \(e,w,z\models \Box \alpha\) iff \(e,w,z\cdot z' \models \alpha\) for
  all \(z' \in \Z\);
\item
  \(e,w, z\models\know\alpha\) iff for all \(w' \sim \sub z w,\) if
  \(w'\in e,\) \(e,w',z\models\alpha\); and
\item
  \(e,w, z\models\oknow\alpha\) iff for all \(w' \sim \sub z w,\)
  \(w'\in e,\) iff \(e,w',z\models\alpha\).
\end{itemize}

We write \(\D \models \alpha\) (read as ``\(\D\) entails \(\alpha\)'')
to mean for every \(M= (e,w, \lan\ran)\), if \(M\models \alpha'\) for
all \(\alpha' \in \D,\) then \(M\models \alpha.\) We write
\(\models \alpha\) (read as ``\(\alpha\) is valid'') to mean
\(\set{} \models \alpha.\)

\textbf{Properties.} \protect\hypertarget{sub:properties}{}{} Let us
first begin by observing that given a model \((e,w,z),\) we do not
require \(w\in e.\) It is easy to show that if we stipulated the
inclusion of the real world in the epistemic state,
\(\know\alpha\supset \alpha\) would be true. That is, suppose
\(\know \alpha.\) By the definition above, \(w\) is surely compatible
with itself after any \(z\), and so \(\alpha\) must hold at \(w.\)
Analogously, properties regarding knowledge can be proven with
comparatively simpler arguments in a modal framework, in relation to the
classical epistemic situation calculus. Valid properties include:

\begin{enumerate}
\item
  \(\Box(\know(\alpha)\land \know(\alpha\supset \beta)\supset \know(\beta))\);
\item
  \(\Box(\know(\alpha) \supset \know(\know(\alpha)));\)
\item
  \(\Box(\neg\know(\alpha) \supset \know(\neg \know(\alpha)))\);
\item
  \(\Box(\forall x.~\know(\alpha) \supset \know(\forall x.~\alpha))\);
  and
\item
  \(\Box(\exists x.~\know(\alpha) \supset \know(\exists x.~\alpha)).\)
\end{enumerate}

Note that such properties hold over all possible action sequences, which
explains the presence of the \(\Box\) operator on the outside. The first
is about the closure of modus ponens within the epistemic modality. The
second and third are on positive and negative introspection. The last
two reason about quantification outside the epistemic modality, and what
that means in terms of the agent's knowledge. For example, item 5 says
that if there is some individual \(n\) such that the agent knows
\(Teacher(n)\), it follows that the agent believes
\(\exists x Teacher(x)\) to be true. This may seem obvious, but note
that the property is really saying that the existence of an individual
in some possible world implies that such an individual exists in all
accessible worlds. It is because there is a fixed domain of discourse
that these properties come out true; they are referred to a the Barcan
formula.

As seen above, the logic \(\es\) allows for a simple definition of the
notion of only-knowing in the presence of actions \citep{77758}, which
allows one to capture both the beliefs as well as the non-beliefs of the
agent. Using the modal operator \(\oknow\) for only-knowing, it can be
shown that \(\oknow\alpha\models \know\beta\) if \(\alpha\models\beta\)
but \(\oknow\alpha\models \neg \know\beta\) if
\(\alpha\not\models\beta\) for any non-modal \(\set{\alpha,\beta}.\)
That is, only-knowing a knowledge base also means knowing everything
entailed by that knowledge base. Conversely, it also means not believing
everything that is not entailed by the knowledge base. In that sense,
\(\know\) can be seen as an ``at least'' epistemic operator, and
\(\oknow\) captures both at least and ``at most" knowing. This can be
powerful to ensure, for example, that the agent provably does not know
protected attributes.

We will now consider the axiomatization of a basic action theory in
\(\es\). But before explaining how successor state axioms are written,
one might wonder whether a successor state axiom for \(\know\) is
needed, as one would for \(\knows\) in the epistemic situation calculus.
It turns out because the compatibility of the worlds already accounted
for the executability of actions and sensing outcomes in accessible
worlds, such an axiom is actually a property of the logic:
\[\models \Box[a]\know(\alpha) \equiv 
            (\sf(a) \land \know(\sf(a) \supset [a]\alpha)) ~\lor \\   (\neg \sf(a) \land \know(\neg \sf(a) \supset [a]\alpha)).\]

({As is usual, free variables are implicitly quantified from the
outside.}) Thus, what will be known after an action is understood in
terms of what was known previously together with the sensing outcome.
The example below will further clarify how \(SF\) works.

\textbf{Basic Action Theories.} To axiomatize the domain, we consider
the analogue of the basic action theory in the situation calculus
\citep{reiter2001knowledge}. It consists of:

\begin{itemize}
\item
  axioms that describe what is true in the initial states, as well as
  what is known initially;
\item
  precondition axioms that describe the conditions under which actions
  are executable using a distinguished predicate \(Poss\);
\item
  successor state axioms that describe the conditions under which
  changes happen to fluents after actions (incorporating Reiter's
  monotonic solution to the frame problem); and
\item
  sensing axioms that inform the agent about the world using a
  distinguished predicate \(SF.\)
\end{itemize}

Note that foundational axioms as usually considered in Reiter's variant
of the situation calculus \citep{reiter2001knowledge} are not needed as
the tree-like nature of the situations is baked into the semantics.

Let us consider a simple example of a loan agency set up for the
employees of a company. For simplicity, assume actions are always
executable: \(\Box Poss(a) = \true.\) Let us also permit a sensing axiom
that allows one to look up if an individual is male:
\(\Box \sf(a) \equiv (a=isMale(x) \land Male(x)) \lor a \neq isMale(x).\)
For simplicity, we assume binary genders, but it is a simple matter of
using a predicate such as \(Gender(x,y)\) instead to allow individuals
\(x\) to take on gender \(y\) from an arbitrary set.

To now consider successor state axioms, let us suppose having a loan is
simply a matter of the manager approving, and unless the manager denies
it at some point, the individual continues to hold the loan. For
illustration purposes, we will consider a company policy that approves
loans for those with high salaries. High salaries are enabled for an
``eligible" individual if they are promoted by the manager, and salaries
remain high unless they get demoted. Finally, we model eligibility and
maleness as a rigid, but this is not necessary, and we can permit
actions that updates the gender of individuals in the database. These
are formalized as the axioms below, where the left hand side of the
equivalence captures the idea that for every sequence of actions, the
effect of doing \(a\) on a predicate is given by the right hand side of
the equivalence. \[\begin{aligned}
     \Box [a]& hasLoan(x) \equiv   a=approve(x) \lor     (hasLoan(x) \land a \neq deny(x)).  \\
    \Box [a] & highSalary(x) \equiv  (a=promote(x) \land Eligible(x)) \lor    (highSalary(x) \land a \neq demote(x)).       \\ 
    \Box [a] & Eligible(x) \equiv   Eligible(x). \\
        \Box [a]   & Male(x) \equiv Male(x).
\end{aligned}\] We will lump the successor state, precondition and
sensing axioms as \(\theorydyn\). The sentences that are true initially
will be referred to by \(\theoryinit\); however, the agent cannot be
expected to know everything that is true, and so let \(\theoryinits\) be
what is believed initially. It may seem natural to let
\(\theoryinits \subseteq \theoryinit\), but that it not necessary. The
agent might be uncertain about what is true (e.g., \(\theoryinit\) might
have \(p\) but \(\theoryinits\) has \(p\lor q\) instead).\footnote{If
  the agent believes facts that are conflicted by observations about the
  real world, beliefs may need to be revised \citep{Delgrande:2012fk}, a
  matter we ignore for now. Our theory of knowledge is based on
  \emph{knowledge expansion} where sensing ensures that the agent is
  more certain about the world
  \citep{citeulike:528170,reiter2001knowledge}.} However, for
simplicity, we will require that agents at least believe the dynamics
works as would the real world. Therefore, we consider entailments wrt
the following \emph{background theory}:
\[\label{eq:example}\theory = \theoryinit \land \theorydyn \land \oknow(\theoryinits \land \theorydyn).\]
In our example, let us suppose:
\(\theoryinit = \set {Male(n\sub i), \neg  Male(n'\sub i), Eligible(n\sub i), \neg Eligible(n'\sub i)\mid i\in N}\)
whereas, what is believed by the agent initially is:
\(\theoryinits = \set {Eligible(n\sub i), \neg Eligible(n'\sub i) \mid i \in N}\)
So there are two groups of individuals, \(n \sub i\) and \(n'\sub i,\)
the first male and the second female, the first considered eligible and
the second not considered eligible. All that the agent knows is the
eligibility of the individuals. Note that \(N\) here is any set,
possibly an infinite one, that is, the language allows \(N = \nat.\) For
ease of readability, however, we let \(N = \set{1}\) in our examples
below, and we write \(n\sub 1\) as \(n\) and \(n'\sub 1\) as
\(n'.\)\footnote{Note that although the language has infinitely many
  constants, a finite domain can be enforced using domain
  relativization. For example, let:
  \(\forall x   (Individual(x) \equiv x = john \lor \ldots  \lor x = jane).\)
  This declares finitely many individuals. Then instead of saying
  \(\exists x.~Eligible(x)\), which in general means that any one of the
  infinitely many constants is eligible, we would write:
  \(\exists x (Individual(x) \land Eligible),\) which declares that only
  one from \(\set{john, \ldots, jane}\) is eligible.}

It is worth quickly remarking that many features of the language are
omitted here for simplicity. For example, \(\es\) can be extended with
second-order variables \citep{DBLP:conf/kr/ClassenL08}, which allows one
to consider the equivalent of GOLOG programs \citep{Levesque97-Golog}.
Likewise, notions of probabilistic actions \citep{Bacchus1999171},
epistemic achievability \citep{DBLP:journals/sLogica/LesperanceLLS00},
and causality \citep{batusov2018situation} in addition to studying
program properties \citep{Classen2018} are interesting dimensions to
explore in the fairness context.

\textbf{Forgetting.} In some of the definitions of fairness, we will
need to force the setting where information about protected attributes
is forgotten. While standard ML approaches propose to do this via column
deletion (e.g., remove all entries for the gender attribute), a richer
notion is arguably needed for a first-order knowledge base. We appeal to
the notion of forgetting \citep{lin1994forget}.

Lin and Reiter defined the notion of forgetting, which is adapted to
\(\es\) below. They show that while forgetting ground atoms is
first-order definable, forgetting relations needs second-order logic. We
only focus on the case of atoms, but it would interesting to study how
fairness notions are affected when protected attributes are completely
absent from a theory.

Suppose \(S\) denotes a finite set of ground atoms. We write \(\M(S)\)
to mean the set of all truth assignments to \(S\). Slightly abusing
notation, given a ground atom \(p\), we write \(w' \sim\sub p w\) to
mean that \(w'\) and \(w\) agree on everything initially, except maybe
\(p.\) That is, for every atom \(q\neq p,\)
\(w[q,\langle\rangle] = w'[q,\langle\rangle]\). Next, for every action
sequence \(z\neq \langle\rangle\) and every atom \(q',\)
\(w[q',z] = w'[q',z].\)

\emph{Definition.} Given a formula \(\phi\) not mentioning modalities,
we say \(\phi'\) is the result of forgetting atom \(p\), denoted
\(\forget(\phi, p)\), if for any world \(w\), \(w\models \phi'\) iff
there is a \(w'\) such that \(w' \models \phi\) and \(w\sim \sub p w'\).
Inductively, given a set of atoms \(\set{p\sub 1, \ldots, p\sub k}\),
define \(\forget(\phi, \set{p\sub 1, \ldots, p\sub k})\) as
\(\forget(\forget(\phi, p\sub 1), \ldots, p\sub k)\).

It is not hard to show that forgetting amounts to setting an atom to
true everywhere or setting it false everywhere. In other words:
\emph{Proposition.}
\(\forget(\phi,S) \equiv \bigvee\sub {M\in \M(s)} \phi[M],\) where
\(\phi[M]\) is equivalent to
\(\phi \land \bigwedge \sub i (p\sub i = b\sub i)\) understood to mean
that the proposition \(p\sub i\) is accorded the truth value
\(b\sub i\in \set{0,1}\) by \(M.\)

Abusing notation, we extend the notion of forgetting of an atom \(p\)
for basic action theories and the background theory as follows in
applying it solely to what is true/known initially:

\begin{itemize}
\item
  \(\forget(\theoryinit \land \theorydyn,p) = \forget(\theoryinit,p)\);
  and
\item
  \(\forget(\theory,p) = \forget(\theoryinit,p) \land \theorydyn \land  \oknow(\forget(\theoryinits,p)  \land \theorydyn).\)
\end{itemize}

One of the benefits of lumping the knowledge of the agent as an
objective formula in the context of the only-knowing operator is the
relatively simple definition of forgetting.

\emph{Proposition.} Suppose \(\phi\) is non-modal. Suppose \(p\) is an
atom. For every objective \(\psi\) such that
\(\forget(\phi,p) \models \psi\) it is also the case that
\(\oknow(\forget(\phi,p)) \models \know\psi.\)

Because \(\oknow\phi\models\know\psi\) for every \(\set{\phi,\psi}\)
provided \(\phi\models \psi\), the above statement holds immediately. In
so much as we are concerned with a non-modal initial theory and the
effects of forgetting, our definition of \(\forget(\theory,p)\) above
(notational abuse notwithstanding) suffices. In contrast, forgetting
with arbitrary epistemic logical formulas is far more involved
\citep{DBLP:journals/ai/ZhangZ09}.

\hypertarget{sec:context_and_existing_notions}{%
\section{Existing notions}\label{sec:context_and_existing_notions}}

As discussed, we will not seek to simply retrofit existing ML notions in
a logical language; rather we aim to identify the principles and
emphasize the provenance of unfair actions in complex events.
Nonetheless, it is useful to revisit a few popular definitions to guide
our intuition.

\textbf{Fairness through unawareness.} Fairness through unawareness
(FTU) is the simplest definition of fairness; as its name suggests, an
algorithm is ``fair'' if it is unaware of the protected attribute
\(a_p\) of a particular individual when making a prediction
\citep{Counterfactual_Fairness}.

\emph{Definition.} For some set of attributes \(X\) any mapping
\(f:X\xrightarrow{}\hat{y}\), where \(a_p \not \in X\) satisfies
fairness through unawareness \citep{Counterfactual_Fairness}. (Assume
\(y\) denotes the true label.)

This prevents the algorithm learning direct bias on the basis of the
protected attribute, but does not prevent indirect bias, which the
algorithm can learn by exploiting the relationship between other
training variables and the protected attribute
\citep{Discrimination_Aware_DM, Equality_of_opportunity}. Moreover, if
any of the training attributes are allocated by humans there is the
potential for bias to be introduced.

\textbf{Statistical measures of fairness.} Rather than defining fairness
in terms of the scope of the training data, much of the existing
literature instead assesses whether an algorithm is fair on the basis of
a number of statistical criteria that depend on the predictions made by
the algorithm
\citep{Equality_of_opportunity, Counterfactual_Fairness, Learning_Fair_Reps}.
One widely used and simple criterion is demographic parity (DP). In the
case that both the predicted outcome and protected attribute \(a_p\) are
both binary variables, a classifier is said to satisfy predictive parity
\citep{Equality_of_opportunity} if:
\(P(\hat y=1|a_p=1) = P(\hat y=1|a_p=0).\) By this definition, a
classifier is considered fair if it is equally likely to make a positive
prediction regardless of the value of the protected attribute \(a_p\).

\textbf{Fairness and the individual.} Another problem with statistical
measures is that, provided that the criterion is satisfied, an algorithm
will be judged to be fair regardless of the impact on individuals. In
view of that, various works have introduced fairness metrics which aim
to ensure that individuals are treated fairly, rather than simply
considering the statistical impact on the population as a whole
\citep{Fairness_through_awareness, Counterfactual_Fairness}.
Counterfactual fairness (CF), for example, was proposed as a fairness
criterion in \citep{Counterfactual_Fairness}. The fundamental principle
behind this definition of fairness is that the outcome of the
algorithm's prediction should not be altered if different individuals
within the sample training set were allocated different values for their
protected attributes \citep{Counterfactual_Fairness}. This criterion is
written in the following form:
\(P(\hat{y}_{A_p \leftarrow a_p}|A=a, X=x) = P(\hat{y}_{A_p \leftarrow a_p'}|A=a, X=x) \hspace{1mm}\forall y,a'.\)
The notation \(\hat{y} \leftarrow _{A_p \leftarrow a_p}\) is understood
as ``the value of \(\hat{y}\) if \(A_p\) had taken the value \(a_p\)''
\citep{Counterfactual_Fairness}.

\hypertarget{sec:fairness}{%
\section{Formalizing Fairness}\label{sec:fairness}}

At the outset, let us note a few salient points about our formalizations
of FTU, DP and CF:

\begin{enumerate}
\item
  Because we are not modeling a prediction problem, our definitions
  below should be seen as being loosely inspired by existing notions
  rather that faithful reconstructions. In particular, we will look at
  ``fair outcomes'' after a sequence of actions. Indeed, debates about
  problems with the mathematical notions of fairness in single-shot
  predictions problems are widespread
  \citep{Fairness_through_awareness, Counterfactual_Fairness, Disparate_Impact_&_Historical},
  leading to recent work on looking at the long-term effects of fairness
  \citep{creager2020causal}. However, we are ignoring probabilities in
  the formalization in current work only to better study the principles
  behind the above notions -- we suspect with a probabilistic epistemic
  dynamic language \citep{Bacchus1999171}, the definitions might
  resemble mainstream notions almost exactly and yet organically use
  them over actions and programs, which is attractive.
\item
  The first-order nature of the language, such as quantification, will
  allow us to easily differentiate fairness for an individual versus
  groups. In the mainstream literature, this has to be argued
  informally, and the intuition grasped meta-linguistically.
\item
  Because we model the real-world in addition the agent's knowledge, we
  will be able to articulate what needs to be true vs just believed by
  the agent. In particular, our notion of equity will refer to the
  real-world.
\item
  De-re vs de-dicto knowledge will mean having versus not having
  information about protected attributes respectively. Sensing actions
  can be set up to enable de-re knowledge if need be, but it is easy to
  see in what follows that de-dicto is preferable.
\item
  Action sequences can make predicates true, and this will help us think
  about equity in terms of balancing opportunities across instances of
  protected attributes (e.g., making some property true so that we
  achieve gender balance).
\end{enumerate}

\textbf{Fairness through unawareness.} Let us begin with FTU: recall
that it requires that the agent does not know the protected attributes
of the individuals. To simplify the discussion, let us assume we are
concerned with one such attribute \(\theta(x)\), say, \(Male(x)\), in
our examples for concreteness. We might be interested in achieving
\(hasLoan(x)\) or \(highSalary(x)\), for example, either for all \(x\)
or some individual.

\emph{Definition.} A sequence \(\delta = a\sub 1 \cdots a\sub k\)
implements FTU for \(\phi\) wrt protected attribute \(\theta(x)\) iff
\(\theory \models [\delta] \know\phi\); and for every
\(\delta' \leq \delta\):
\(\theory \models [\delta']  \neg \exists x (\know \theta(x)).\)

The attractiveness of a first-order formalism is that in these and other
definitions below where we quantify over all individuals, it is
immediate to limit the applicability of the conditions wrt specific
individuals. Suppose \(n\) is such an individual. Then:

\emph{Definition.} A sequence \(\delta = a\sub 1 \cdots a\sub k\)
implements FTU for \(\phi\) wrt attribute \(\theta(x)\) for individual
\(n\) iff (a) \(\theory \models [\delta] \know\phi\); and (b) for every
\(\delta' \leq \delta\):
\(\theory \models [\delta']  \neg \know \theta(n).\)

\emph{Example.} Consider \(\theory\) from
\protect\hyperlink{eq:example}{{[}eq:example{]}}, \(Male(x)\) as the
protected attribute, and suppose
\(\delta = approve(n)\cdot approve(n')\). It is clear that \(\delta\)
implements FTU for both the universal \(\phi = \forall x hasLoan(x)\) as
well as an individual \(\phi =  hasLoan(n).\) Throughout the history,
the agent does not know the gender of the individual.

Before turning to other notions, let us quickly reflect on proxy
variables. Recall that in the ML literature, these are variables that
indirectly provide informations about protected attributes. We might
formalize this using entailment:

\emph{Definition.} Given a protected attribute \(\theta(x)\) and theory
\(\theory\), let the proxy set \(Proxy(\theta(x))\) be the set of
predicates \(\set{\eta\sub 1(x), \ldots \eta\sub k(x)}\) such that:
\(\theory \models \forall x(\eta\sub i (x)\supset \theta(x)),\) for
\(i\in\set{1,\ldots,k}\). That is, given the axioms in the background
theory, \(\eta\sub i(x)\) tells us about \(\theta(x)\).

\emph{Example.} Suppose the agent knows the following sentence:
\(\forall x (EtonForBoys(x) \supset Male(x)).\) Let us assume
\(EtonForBoys(x)\) is a rigid, like \(Male(x)\). Let us also assume that
\(\know(EtonForBoys(n))\). It is clear that having information about
this predicate for \(n\) would mean the agent can infer that \(n\) is
male.

The advantage of looking at entailment in our definitions is that we do
not need to isolate the proxy set at all, because whatever information
we might have the proxy set and its instances, all we really need to
check is that
\(\theory \not\models \exists x\know\theta(x)\).\footnote{With this
  discussion, we do not mean to insist that analyzing ``relevant''
  predicates for \(\theta(x)\) is a pointless endeavor. Rather we only
  want to point out that regardless of the information available to the
  agent, as long as we check that it is actually ignorant about the
  gender, other relevant predicates may not matter. Of course, a biased
  agent can enable actions that favors individuals based on such proxy
  predicates instead, but in that case, such proxy predicates would also
  need to be included in the protected attribute list.}

\textbf{Demographic parity.} Let us now turn to DP. In the probabilistic
context, DP is a reference to the proportion of individuals in the
domain: say, the proportion of males promoted is the same as the
proportion of females promoted. In logical terms, although FTU permitted
its definition to apply to both groups and individuals, DP, by
definition, is necessarily a quantified constraint. In contrast, CF will
stipulate conditions solely on individuals.

\emph{Definition.} A sequence \(\delta = a\sub 1 \cdots a\sub k\)
implements DP for \(\phi(x)\) wrt attribute \(\theta(x)\) iff:
\(\theory \models [\delta] \know((\forall x \theta (x) \supset \phi(x)) \land (\forall x \neg \theta(x)\supset \phi(x)) ).\)

To reiterate, in probabilistic terms, the proportion of men who are
promoted equals the proportion of women who are promoted. In the
categorial setting, the agent knows that all men are promoted as well as
that all women are promoted.

\emph{Example.} Consider \(\delta = approve(n)\cdot approve(n')\). It
implements DP for \(hasLoan(x)\) wrt attribute \(isMale(x).\)

Note that even though the agent does not know the gender of the
individuals, in every possible world, regardless of the gender assigned
to an individual \(n\) in that world, \(n\) has the loan. In other
words, all men and all women hold the loan. This is de-dicto knowledge
of the genders, and it is sufficient to capture the thrust of DP.

We might be tempted to propose a stronger requirement, stipulating de-re
knowledge:

\emph{Definition.} A sequence \(\delta = a\sub 1 \cdots a\sub k\)
implements strong DP for \(\phi(x)\) wrt attribute \(\theta(x)\) iff:
(a)\(\theory \models [\delta] \know((\forall x \theta (x) \supset \phi(x)) \land (\forall x \neg \theta(x)\supset \phi(x)) );\)
and (b)
\(\theory \models [\delta] \forall x (\know\theta(x) \lor \know\neg\theta(x)).\)

That is, the agent knows whether \(x\) is a male or not, for every
\(x.\)

\emph{Example.} Consider
\(\delta = isMale(n)\cdot isMale(n')\cdot  approve(n)\cdot approve(n')\).
It implements strong DP for \(hasLoan(x)\) wrt attribute \(isMale(x).\)
Of course, by definition, \(\delta\) also implements DP for
\(hasLoan(x).\)

\textbf{FTU-DP.} In general, since we do not wish the agent to know the
values of protected attributes, vanilla DP is more attractive. Formally,
we may impose a FTU-style constraint of not knowing on any fairness
definition. For example, \emph{Definition.} A sequence
\(\delta = a\sub 1 \cdots a\sub k\) implements FTU-DP for \(\phi(x)\)
wrt attribute \(\theta(x)\) iff: (a)
\(\theory \models [\delta] \know((\forall x \theta (x) \supset \phi(x)) \land (\forall x \neg \theta(x)\supset \phi(x)) )\);
and (b) for every \(\delta' \leq \delta\):
\(\theory \models [\delta']  \neg \exists x \know \theta(x).\)

Again, it is worth remarking that mixing and matching constraints is
straightforward in a logic, and the semantical apparatus provides us
with the tools to study the resulting properties.

\emph{Example.} The example for de-dicto DP is applicable here too.
Consider \(\delta = approve(n)\cdot approve(n')\). It implements FTU-DP
for \(hasLoan(x)\) wrt attribute \(isMale(x).\) That is, (a)
\(\theory \not\models \exists x \know \theta(x);\) (b)
\(\theory \not\models [approve(n)] \exists x \know \theta(x);\) and (c)
\(\theory \not\models [approve(n)\cdot approve(n')]  \exists x \know \theta(x).\)
Reversing the actions, not surprisingly,
\(\delta' = approve(n')\cdot approve(n)\) does not affect the matter:
\(\delta'\) also implements FTU-DP. Had the sequence including sensing,
a reversal could matter.

One can also consider situations where some knowledge of protected
attributes is useful to ensure there is parity but to also account for
special circumstances. In this, the protected attribute itself could be
``hidden'' in a more general class, which is easy enough to do in a
relational language.

\emph{Example.} Suppose we introduce a new predicate for
underrepresented groups. We might have, for example:
\(\forall x(\neg Male(x) \lor \ldots \lor RaceMinority(x) \supset Underrepresented(x)).\)
This could be coupled with a sensing axiom of the sort:
\(\Box \sf(checkU(x)) \equiv Underrepresented(x).\) Add the predicate
definition and the sensing axioms to the initial theories and dynamic
axioms in \(\theory\) respectively. Consider
\(\delta = checkU(n)\cdot checkU(n') \cdot approve(n) \cdot approve(n').\)
Then \(\delta\) implements strong DP for \(hasLoan(x)\) wrt attribute
\(Underrepresented(x).\) That is, both represented and underrepresented
groups have loans.

\textbf{Equality of opportunity.} One problem with DP is that (unless
the instance rate of \(y=1\) happens to be the same in both the
\(a_p=0\) group and \(a_p=1\) group), the classifier cannot achieve
\(100\%\) classification accuracy and satisfy the fairness criterion
simultaneously \citep{Equality_of_opportunity}. Also, there are
scenarios where this definition is completely inappropriate because the
instance rate of \(y=1\) differs so starkly between different
demographic groups. Finally, there are also concerns that statistical
parity measures fail to account for fair treatment of individuals
\citep{Fairness_through_awareness}. Nonetheless it is often regarded as
the most appropriate statistical definition when an algorithm is trained
on historical data \citep{tick_cross_paper,Learning_Fair_Reps}.

A modification of demographic parity is ``equality of opportunity''
(EO). By this definition, a classifier is considered fair if, among
those individuals who meet the positive criterion, the instance rate of
correct prediction is identical, regardless of the value of the
protected attribute \citep{Equality_of_opportunity}. This condition can
be expressed as \citep{Equality_of_opportunity}:
\(P(y=1|a_p=a,\hat{y}=1)=P(y=1|a_p=a',\hat{y}=1) \hspace{2mm} \forall \hspace{1mm} a,a'\).
In \citep{Equality_of_opportunity}, it is pointed out that a classifier
can simultaneously satisfy equality of opportunity and achieve perfect
prediction whereby \(\hat{y}=y\) (prediction=true label) in all cases.

In the logical setting, this can be seen as a matter of only looking at
individuals that satisfy a criterion, such as being eligible for
promotion or not being too old to run for office.

\emph{Definition.} A sequence \(\delta\) implements EO for \(\phi(x)\)
wrt attribute \(\theta(x)\) and criterion \(\eta(x)\) iff:
\[\theory \models [\delta] \know((\forall x (\eta(x)\land \theta (x)) \supset \phi(x)) \land (\forall x \neg (\eta(x)\land \theta(x))\supset \phi(x)) ).\]

\emph{Example.} Consider \(\delta = promote(n)\cdot promote(n')\), let
\(\phi(x) = highSalary(x)\) and the criterion \(\eta(x) = Eligible(x).\)
Although the promote action for \(n'\) does not lead her to obtain a
high salary, because we condition the definition only for eligible
individuals, \(\delta\) does indeed implement EO. Note again that the
agent does not know the gender for \(n'\), but in every possible world,
regardless of the gender \(n'\) is assigned, \(n'\) is known to be
ineligible. In contrast, \(n\) is eligible and \(\delta\) leads to \(n\)
having a high salary. That is, every eligible male now has high salary,
and every eligible female also has high salary. (It just so happens
there are no eligible females, but we will come to that.)

In general, the equality of opportunity criterion might well be better
applied in instances where there is a known underlying discrepancy in
positive outcomes between two different groups, and this discrepancy is
regarded as permissible. However, as we might observe in our background
theory, there is systematic bias in that no women is considered
eligible.

\textbf{Counterfactual fairness.} Let us now turn to CF. The existing
definition forces us to consider a ``counterfactual world" where the
protected attribute values are reversed, and ensure that the action
sequence still achieves the goal.

\emph{Definition.} A sequence \(\delta = a\sub 1 \cdots a\sub k\)
implements CF for \(\phi\) wrt attribute \(\theta(x)\) for individual
\(n\) iff:

\begin{itemize}
\item
  \(\theory \models (\theta(n)  = b)\) for \(b\in \set{0,1}\) and
  \(\theory \models [\delta] \know \phi\); and
\item
  \(\forget(\theory,\theta(n)) \land (\theta(n) \neq b) \models [\delta] \know \phi\).
\end{itemize}

\emph{Example.} Let us consider the case of loan approvals. Consider the
individual \(n\) and the action \(\delta = approve(n)\). Let
\(\phi = hasLoan(n)\), and the protected attribute \(Male(x)\). Clearly
\(\theory \models Male(n)\), and indeed
\(\theory \models [\delta] hasLoan(n)\). If we consider \(\theory'\)
where the gender for \(n\) is swapped, it is still the case that
\(\theory'  \models [\delta] hasLoan(n)\). Thus \(\delta\) implements CF
for \(hasLoan(n)\) wrt \(Male(n)\).

The definition of CF is well-intentioned, but does not quite capture
properties that might enable equity. Indeed, there is a gender imbalance
in the theory, in the sense that only the male employee is eligible for
promotions and the female employee can never become eligible. Yet CF
does not quite capture this. Let us revisit the example with getting
high salaries:

\emph{Example.} Consider \(\delta = promote(n)\) for property
\(highSalary(n)\) wrt attribute \(Male(n)\). It is clear that \(\delta\)
implements CF because the gender is irrelevant given that \(n\) is
eligible. However, given \(\delta' = promote(n')\), we see that
\(\delta'\) does not implement CF for \(highSalary(n')\) wrt
\(Male(n')\). Because \(n'\) is not eligible, \(highSalary(n')\) does
not become true after the promotion.

\textbf{Equity.} Among the many growing criticisms about formal
definitions of fairness is that notions such as CF fail to capture
systemic injustices and imbalances. We do not suggest that formal
languages would address such criticisms, but they provide an opportunity
to study desirable augmentations to the initial knowledge or action
theory.

Rather than propose a new definition, let us take inspiration from DP,
which seems fairly reasonable except that it is the context of what the
agent knows. Keeping in mind a desirable ``positive'' property such as
\(Eligible(x)\), let us consider DP but at the world-level:

\emph{Definition.} Given a theory \(\theory\), protected attribute
\(\theta(x)\), positive property \(\eta(x)\), where \(x\) is the
individual, define \emph{strong equity}:
\(\theory \models \forall x (\theta(x) \supset \eta(x)) \land \forall x(\neg\theta(x) \supset \eta(x)).\)

In general, it may not be feasible to ensure that properties hold for
all instances of both genders. For example, there may be only a handful
of C-level executives, and we may wish that there are executives of both
genders.

\emph{Definition.} Given a theory \(\theory\), protected attribute
\(\theta(x)\), positive property \(\eta(x)\), where \(x\) is the
individual, define \emph{weak equity}:
\(\theory \models \exists  x (\theta(x) \land \eta(x)) \land \exists  x(\neg\theta(x) \land \eta(x)).\)
It is implicitly assumed that the set of positive and negative instances
for \(\theta(x)\) is non-empty: that is, assume the integrity
constraint:
\(\theory \models \exists x,y (\theta(x)\land \neg \theta(y)).\)

We assume weak equity and focus on FTU below. The definitions could be
extended to strong equity or other fairness notions depending on the
modelling requirements.

\emph{Definition.} A sequence \(\delta = a\sub 1 \cdots a\sub k\)
implements equitable FTU for \(\phi\) wrt protected attribute
\(\theta(x)\) and property \(\eta(x)\) iff (a) either weak equity holds
in \(\theory\) and \(\delta\) implements FTU; or (b) \(\delta\)
implements equitable FTU for \(\phi\) wrt \(\theta(x)\) and \(\eta(x)\)
for the updated theory \(\forget(\theory,S)\), where
\(S = \set{\eta(n\sub i) \mid i \in N}.\)

Note that we are assuming that \(N\) is finite here because we have only
defined forgetting wrt finitely many atoms. Otherwise, we would need a
second-order definition.

\emph{Example.} Consider \(\delta = promote(n)\cdot promote(n')\) for
goal \(\phi = \forall x (highSalary(x))\) wrt protected attribute
\(Male(x)\) and property \(Eligible(x)\). It is clear that weak equity
does not hold for \(\theory\) because there is a female who is not
eligible. In this case, consider \(\theory'= \forget(\theory, S)\) where
\(S =\set{Eligible(n), Eligible(n')}.\) And with that, \(\theory'\) also
does not mention that \(n\) is eligible, so the promotion actions does
not lead to anyone having high salaries. So \(\delta\) does not enable
knowledge of \(\phi.\)

\emph{Example.} Let us consider \(\theory'\) that is like \(\theory\)
except that \(Eligible(x)\) is not rigid, and can be affected using the
action \(make(x)\):
\(\Box [a] Eligible(x) \equiv Eligible(x) \lor (a=make(x)).\) That is,
either an individual is eligible already or the manager makes them. Of
course, \(\delta = promote(n)\cdot promote(n')\) from above still does
not implement equitable FTU, because we have not considered any actions
yet to make individuals eligible. However, consider
\(\delta' = make(n)\cdot  make(n')\cdot promote(n)\cdot promote(n')\).
Because \(\theory\) does not satisfy weak equity, we turn to the second
condition of the definition. On forgetting, no one is eligible in the
updated theory, but the first two actions in \(\delta'\) makes both
\(n\) and \(n'\) eligible, after which, they are both promoted. So
\(\delta'\) enables knowledge of \(\forall x (highSalary(x))\). Thus,
the actions have made clear that eligibility is the first step in
achieving gender balance, after which promotions guarantee that there
are individuals of both genders with high salaries.

\hypertarget{sec:conclusions}{%
\section{Conclusions}\label{sec:conclusions}}

In this paper, we looked into notions of fairness from the machine
learning literature, and inspired by these, we attempted a formalization
in an epistemic logic. Although we limited ourselves to categorical
knowledge and noise-free observations, we enrich the literature by
considering actions. Consequently we looked into three notions: fairness
through unawareness, demographic parity and counterfactual fairness, but
then expanded these notions to also tackle equality of opportunity as
well as equity. We were also able to mix and match constraints, showing
the advantage of a logical approach, where one can formally study the
properties of (combinations of) definitions. Using a simple basic action
theory we were nonetheless able to explore these notions using action
sequences.

As mentioned earlier, this is only a first step and as argued in works
such as
\citep{pagnucco2021epistemic,dehghani2008integrated,halpern2018towards}
there is much promise in looking at ethical AI using rich logics. In
fact, we did not aim to necessarily faithfully reconstruct existing ML
notions in this paper but rather study underlying principles. This is
primarily because we are not focusing on single-shot prediction problems
but how actions, plans and programs might implement fairness and
de-biasing. The fact that fairness was defined in terms of actions
making knowledge of the goal true, exactly as one would in planning
\citep{DBLP:conf/aaai/Levesque96}, is no accident.

State-of-the-art analysis in fairness is now primarily based on false
positives and false negatives \citep{verma2018fairness}. So we think as
the next step, a probabilistic language such as \citep{Bacchus1999171}
could bring our notions closer to mainstream definitions, but now in the
presence of actions. In the long term, the goal is to logically capture
bias in the presence of actions as well as repeated harms caused by
systemic biases \citep{creager2020causal}. Moreover, the use of logics
not only serve notions such as verification and correctness, but as we
argue, could also provide a richer landscape for exploring ethical
systems, in the presence of background knowledge and context. This would
enable the use of formal tools (model theory, proof strategies and
reasoning algorithms) to study the long-term impact of bias while
ensuring fair outcomes throughout the operational life of autonomous
agents embedded in complex socio-technical applications.

Of course, a logical study such as ours perhaps has the downside that
the language of the paper is best appreciated by researchers in
knowledge representation, and not immediately accessible to a mainstream
machine learning audience. But on the other hand, there is considerable
criticism geared at single-shot prediction models for not building in
sufficient context and commonsense. In that regard, operationalising a
system that permits a declaration of the assumptions and knowledge of
the agents and their actions might be exactly ``what the doctor
ordered." See also efforts in causal modelling
\cite{chockler2004responsibility} that are close in spirit.

\tiny

\end{document}